\documentclass{llncs}
\usepackage[pdftex]{graphicx}
\usepackage{url}
\usepackage{mathtools}

\begin{document}
\title{\textbf{Augmented Utilitarianism for AGI Safety }}
\author{Nadisha-Marie Aliman\inst{1} \and Leon Kester\inst{2}}
\institute{Utrecht University, Utrecht, Netherlands \and TNO Netherlands, The Hague, Netherlands }
\maketitle

\begin{abstract}
	
In the light of ongoing progresses of research on artificial intelligent systems exhibiting a steadily increasing problem-solving ability, the identification of practicable solutions to the value alignment problem in AGI Safety is becoming a matter of urgency. In this context, one preeminent challenge that has been addressed by multiple researchers is the adequate formulation of utility functions or equivalents reliably capturing human ethical conceptions. However, the specification of suitable utility functions harbors the risk of ``perverse instantiation" for which no final consensus on responsible proactive countermeasures has been achieved so far. Amidst this background, we propose a novel socio-technological ethical framework denoted \textit{Augmented Utilitarianism} which directly alleviates the perverse instantiation problem. We elaborate on how augmented by AI and more generally science and technology, it might allow a society to craft and update ethical utility functions while jointly undergoing a dynamical ethical enhancement. Further, we elucidate the need to consider embodied simulations in the design of utility functions for AGIs aligned with human values. Finally, we discuss future prospects regarding the usage of the presented scientifically grounded ethical framework and mention possible challenges.
	
\end{abstract}

\section{Motivation }
 \label{sec:intro}
The problem of unambiguously specifying human goals for advanced AI systems such that these systems once deployed, do not violate the implicitly underlying intended human conceptions by pursuing unforeseen solutions, has been referred to as ``literalness"\cite{meuhlhauser2012intelligence,yampolskiy2015artificial} or also ``perverse instantiation" \cite{Bostrom:2014:SPD:2678074,y} problem. A higher problem solving ability does not necessarily entail the integration of the contextual knowledge required from an advanced AI in order to accurately interpret human ethical conceptions. Therefore, it is of great importance from the perspective of A(G)I Safety and A(G)I Ethics to a priori consider this crucial issue when crafting utility functions for intelligent systems that would operate based on the human goals these functions encode. Recently, a novel type of such quantitative utility functions denoted \textit{ethical goal functions}~\cite{Delphi,werkhoven2018telling} has been introduced as critical tool for a society to achieve a meaningful control of autonomous intelligent systems aligned with human ethical values. Departing from this, we show why in order to design ethical goal functions and avoid perverse instantiation scenarios, one needs a novel type of ethical framework for the utility elicitation on whose basis these functions are crafted. We introduce a new to be described socio-technological ethical framework denoted \textit{Augmented Utilitarianism} (which we abbreviate with AU in the following).  \par While multiple methods have been suggested as moral theory approaches to achieve ethical objective functions for AGIs~\cite{everitt2018towards,ijcai2018-768} (including classical ethical frameworks like consequentialism or encompassing methods based on uncertain objectives and moral uncertainty~\cite{bogosian2017implementation,Eckersley2018ImpossibilityAU}), most approaches do not provide a fundamental solution to the underlying problem which wrongly appears to be solely of philosophical nature. According to Goertzel~\cite{goertzel2015superintelligence}, \textit{``pithy summaries of complex human values evoke their commonly accepted meanings only within the human cultural context"}. More generally, we argue that in order to craft utility functions that should not lead to a behavior of advanced AI systems violating human ethical intuitions, one has to scientifically consider relevant contextual and embodied information. Moreover, it could be highly valuable to take into account human biases and constraints that obstruct ethical decision-making and attempt to remediate resulting detrimental effects using science and technology. In contrast to the AU approach we will present, most currently known moral theories and classical ethical frameworks considered for advanced AI systems do not integrate these decisive elements and might therefore riskily not exhibit a sufficient safety level with regard to perverse instantiation.

\section{Deconstructing Perverse Instantiation}
\label{sec:dis}

%

In the following, we enumerate (using the generic notation $ <FinalGoal>$ $:$ $ <PerverseInstantiation>$) a few conceivable perverse instantion scenarios that have been formulated in the past:
\begin{enumerate}
	\item ``Make us smile" : ``Paralyze human facial musculatures into constant beaming smiles" (example by Bostrom ~\cite{Bostrom:2014:SPD:2678074})
	\item ``Make us happy" : ``Implant electrodes into the pleasure centers of our brains" (example by Bostrom ~\cite{Bostrom:2014:SPD:2678074})
	\item ``Making all people happy" : ``Killing all people [...] as with zero people around all of them are happy" (example by Yampolskiy~\cite{yampolskiy2015artificial})
	\item ``Making all people happy" : ``Forced lobotomies for every man, woman and child [...]" (example by Yampolskiy~\cite{yampolskiy2015artificial})

\end{enumerate} 
From our view, one could extract the following two types of failures out of the specified perverse instantiations: misspecification of final goal criteria and the so called \textit{perspectival fallacy of utility assignment}~\cite{Delphi} which will become apparant in our explanation. First, one could argue that already the proposed formulations regarding the criteria of the final goal do not optimally capture the nature of the intended sense from a scientific perspective which might have finally misguided the AI. While the concept of happiness certainly represents a highly ambiguous construct, modern research in the field of positive psychology~\cite{lopez2018positive,peterson2006primer,seligman2014positive}, hedonic psychology~\cite{kahneman1999well} and further research areas offers a scientific basis to assess what it means for human entities. For instance, one might come to the conclusion that a highly desirable final goal of humanity for a superintelligence rather represents a concept which is close to the notion of ``well-being". In psychology, well-being has been among others described as a construct consisting of five measurable elements: positive emotions, engagement, relationships, meaning and achievement (PERMA)~\cite{seligman2012flourish}. Another known psychological measure for well-being is subjective well-being~\cite{lyubomirsky2001some} (SWB) which is composed of frequent positive affect, infrequent negative affect and life  satisfaction~\cite{busseri2011review,diener2000subjective}. In both cases, happiness only represents a subelement of the respective well-being construct. Similarly, as stated by Diener and Bieswas-Diener~\cite{diener2011happiness}, \textit{``happiness alone is not enough; people need to be happy for the right reasons"}. Coming back to the provided examples for perverse instantiation, in the cases 1, 2 and 4, it is implausible that a pluralistic criteria of well-being like PERMA would have been met.\par

Second, it is however important to note that even if the final goal would have been specified in a way reflecting psychological insights, a perverse instantiation cannot necessarily be precluded without more ado. By way of illustration, we correspondingly reformulate the example 3 within a new type of perverse instantiation and provide an additional example. We thereby use the term ``flourish" to allude to the achievement of a high level of well-being in line with a psychological understanding of the concept as exemplified in the last paragraph. 
\begin{enumerate}
\setcounter{enumi}{4}
	\item Make all people flourish : Killing all people
	\item Make all people flourish : Initiate a secret genocide until the few uninformed people left in future generations all flourish
\setcounter{enumi}{0}	
\end{enumerate} 
Despite a suitable final goal, value alignment is not succesful in 5 and 6 because the underlying assignment of utility seems to be based on a detached modus operandi in which the effects of scenarios on the \textit{own} current mental states of the people generating this function are ignored. Thereby, it is assumed that during utility assignment, the involved people are considered as remote observers, while at the same time one inherently takes their perspective while referring to this mapping with the emotionally connoted description of a \textit{perverse} instantiation. This type of detached design of utility functions ignoring i.a.\ affective and emotional parameters of the own mental state has been described as being subject to the perspectival fallacy of utility assignment~\cite{Delphi}. Although most people would currently dislike all provided examples 1-6, the aggregated mental states of their current selves seem not to be reflected within the utility function of the AI which instead considered a synthetic detached measure only related to their future selves or/and future living people. In the next paragraph, we briefly introduce a known problem in population ethics  that exhibits similar patterns and which might be of interest for the design of utility functions for advanced AI systems in certain safety-relevant application areas~\cite{Eckersley2018ImpossibilityAU}. \par 
Population ethics~\cite{greaves2017population} is an issue in philosophy concerning decision-making that potentially leads to populations with varying numbers or/and identities of their members. One interesting element of a population ethics theory is the derived population axiology which represents the total order of different population states according to their ethical desirability. As an example, consider the choice of either perform a policy measure that leads to a population A of ca. 10 billion members and a very high positive welfare or to rather prefer another policy measure leading to a population Z of ca. 10.000 billion members and a much lower only barely acceptable (but still positive) welfare. Intuitively, most people would rank the policy measure leading to population A as higher than the one leading to population Z. 
However, given the population axiology of total utilitarianism~\cite{greaves2017population},
Z might well be ranked higher than A 
if the number of people multiplied by their welfare is bigger for population Z in comparison to population A. This type of violation of human ethical intuitions when applying total utilitarianism to population ethics has been termed ``Repugnant Conclusion" by Derik Parfit~\cite{parfit1984reasons}.
  In this context, Arrhenius~\cite{arrhenius2000impossibility} proved in one of his impossibility theorems that no population axiology\setcounter{footnote}{0}\footnote {Importantly, this also applies to non-consequentialist frameworks such as deontological ethics~\cite{greaves2017population}.} can be formulated that concurrently satisfies a certain number of ethical desiderata.\par
  
However, as shown by Aliman and Kester~\cite{Delphi}, this type of impossibility theorem does not apply to population axiologies that take the mental states of those attempting to craft the total orders during utility elicitation into account. Similarly to the perverse instantiation examples 1-6, the application of e.g.\ total utilitarianism to the described scenario is subject to the perspectival fallacy of utility assignment. As in the case of these perverse instantiations, the fact that most people consider the scenario involving population Z as \textit{repugnant} is not reflected in the utility function which only includes a detached measure of the well-being of future people. In practice, how humans rate the ethical desirability of for instance a policy measure leading to a certain population, is dependent on the effect the mental simulation of the associated scenario has on their corresponding mental states which inherently encode e.g.\ societal, cultural and temporal information. For instance, from the perspective of a current population $ Z_0 $ being similar to population Z both with regard to number of people and welfare level, it might instead be ``repugnant" to prefer the policy measure leading to population A~\cite{Delphi}. The reason being that the scenario leading from $ Z_0 $ to A might have included a dying out or even a genocide. The lack of the required contextual information in consequentialist frameworks (such as utilitarianism) has implications for AIs and AGIs that are implemented in the form of expected utility maximizers mostly operating in a consequentialist fashion.

\section{Augmenting Utilitarianism  }
\label{sec:eth}
In the light of the above, it appears vital to refine classical utilitarianism (CU) if one intends to utilize it as basis for utility functions that do not lead to perverse instantiation scenarios. However, as opposed to classical ethical frameworks, AU does not represent a normative theory aimed at specifying what humans\textit{ ought to do}. In fact, its necessity arises directly from a technical requirement for the meaningful control of artificial intelligent systems equipped with a utility function. Since the perverse instantiation problem represents a significant constraint to the design of ethical goal functions, a novel tailored ethical framework able to alleviate issues related to both misspecification of final goal criteria and perspectival fallacy of utility assignment emerges as exigency. With this in mind, AU is formulated as a non-normative ethical framework for AGI Safety which can be augmented by the use of science and technology in order to facilitate a dynamical societal process of crafting and updating ethical goal functions. Instead of specifying what an agent ought to do, AU helps to identify what the current society \textit{should want} an (artificial or human) agent to do if this society wants to maximize expected utility. In this connection, utility could ideally represent a generic scientifically grounded (possibly aggregated) measure capturing one or more ethically desirable final goal(s) as defined by society itself. In the following, we describe by what type of components AU could augment CU:

\begin{itemize}

	\item \textit{Scientific grounding of utility:} 
According to Jeremy Bentham~\cite{Bentham1780-BENITT}, the founder of CU \textit{"by the principle of utility is meant that principle which approves or disapproves of every action whatsoever according to the tendency it appears to have to augment or diminish the happiness of the party whose interest is in question"}. For AU, one could for instance reformulate the principle of utility by substituting ``happiness" with a generic scientific measure for one or more final goal(s). In the context of crafting ethical goal functions, the party whose interest is in question is society. Further, a crucial difference between CU and AU is that in order to assess the tendency an action has to augment or diminish the chosen ethical measure, AU considers more than just the outcome of that action as used in the classical sense, since AU presupposes the \textit{mental-state-dependency}~\cite{Delphi} of utility as will be expounded in the next subitem. With this application-oriented view, one could then argue that what society should ideally want an agent to do are actions that are conformable to this modified mental-state-dependent principle of utility. In this paper, we exemplarily consider well-being as reasonable high level final goal candidate which is e.g.\ already reflected in the UN Sustainable Developmental Goals (SDGs)~\cite{zie} and is in the spirit of positive computing~\cite{calvo2014positive}. Besides SWB and PERMA, multiple measures of well-being exist in psychology with focus on different well-being factors. For instance, the concept of objective happiness~\cite{kahneman1999well} has been proposed by Kahneman. Well-being has moreover been linked to the hierarchy of needs of Abraham Maslow which he extended to contain self-transcendence at the highest level on top of self-actualization in his later years~\cite{kaufman2018self,koltko2006rediscovering,maslow1971farther}. (Recently, related AI research aiming at inducing self-transcendent states for users has been considered by among others Mossbridge and Goertzel~\cite{mossbridge2018emotionally}.) For a review on relevant well-being factors that might be pivotal for a dedicated positive computing, see Calvo and Peters~\cite{busseri2011review}. 
	\item \textit{Mental-state-dependency:} As adumbrated in the last section, human ethical evaluation of an outcome of an action is related to their mental states which take into account the simulation that led to this outcome. The mental phenomenon of actively simulating different alternative scenarios (including anticipatory emotions~\cite{baucells2016temporal}) has been termed conceptual consumption~\cite{gilbert2007prospection} and plays a role in decision-making. Similarly, according to Johnson~\cite{johnson1994moral} \textit{``moral deliberation is a process of cognitive conative affective simulation"}. Moreover, it has been shown that for diverse economical and societal contexts, people do not only value the outcome of actions but also assign a well-being relevant \textit{procedural utility}~\cite{frey2001beyond,kaminitz2019contemporary} to the policy that led to these outcomes. In light of this, AU assigns utility at a higher abstraction level by -- simply put -- considering the underlying state transition (from starting state over action to outcome) instead of the outcome alone as in classical consequential frameworks like CU. 
	 Furthermore, according to constructionist approaches in neuroscience~\cite{barrett2017theory}, the brain constructs mental states based on \textit{``sensations from the world, sensations from the body, and prior experience"}~\cite{oosterwijk2012states}. Hence, it is conceivable that ethical judgements might vary from person to person and from society to society with respect to multiple parameters of different nature including e.g.\ psychological, biographical, cultural, temporal, social and physiological information. Likewise, the recent Moral Machine experiment studying human ethical conceptions on trolley case scenarios with i.a.\ autonomous vehicles showed \textit{``substantial cultural variations"} in the exhibited moral preferences~\cite{awad2018moral}. Ethical frameworks for AGI utility functions that disregard the mental-state-dependency may more likely lead to perverse instantiations, since they ignore what we call the \textit{embodied nature of ethical total orders} which AU\footnote{Thereby, it is important to note that AU is not conceived as an attempt to ``consequentialise"~\cite{peterson2010royal} multiple classical ethical frameworks} attempts to consider. 
\item \textit{Debiasing of utility assignment:} One might regard decision utility based on observed choices (as exhibited e.g.\ in big data) as sufficient utility source for a possible instantiation of AU if one assumes that humans are rational agents that already act as to optimize what increases their well-being. However, utility as measured from this third-person perspective might not capture the actual experienced utility from a first-person perspective due to multiple human cognitive biases~\cite{berridge2014experienced,kahneman1997back}. Since it is impossible to directly extract the instant utility (the basic building block of experienced utility~\cite{kahneman1997back}) of future outcomes to craft ethical goal functions, AU could -- in its most basic implementation -- rely on predicted utility which represents the belief on the future experienced utility people would assign to a given scenario from a first-person perspective. However, the mental simulations on whose basis predicted utility is extracted are still distorted among others due to the fact that humans fail to accurately predict their appreciation of future scenarios~\cite{kahneman1997back}. Therefore, it has been suggested by Aliman and Kester~\cite{Delphi} to augment the utility elicitation process by the utilization of technologies like virtual reality (VR) and augmented reality (AR) within a simulation environment in order to be able to get access to a less biased \textit{artificially simulated future instant utility}. Analogous to the AI-aided policy-by-simulation approach~\cite{werkhoven2018telling}, this technique might offer a powerful preemptive tool for AGI Safety in an AU framework. 
Overall, the experience of possible future world scenarios might improve the quality of utility assignment while having the potential to yield an ethical enhancement for one thing due to the debiased view on the future and secondly, for instance due to beneficial effects that immersive technologies might have on prosocial behavior including certain forms of empathy~\cite{calvo2014positive,van2018virtual}. (Interestingly, the experience of individualized and tailored simulations itself might provide an alternative simulation-based solution to the value alignment problem as specified by Yampolskiy~\cite{yampolskiy2019personal}.)

	\item \textit{Self-reflexivity:} As opposed to CU, AU is intended as a self-reflexive ethical framework which augments itself. Due to the mental-state-dependency it incorporates and the associated embodied nature of ethical total orders, it might even be necessary to craft new ethical goal functions within a so-called socio-technological feedback-loop~\cite{Delphi}. In doing so, ongoing technological progresses might help to augment the debiasing of utility assignment while novel scientific insights might facilitate to filter out the most sophisticated measure for the ethically desired form of utility given the current state of society. Advances in A(G)I development itself leading to a higher problem solving ability might further boost AU with an improved predictability of future outcomes leading to more precise ethical goal functions. Given its generic nature, what humans should want an agent to do might thereby vary qualitatively in an AU framework, since quantitatively specifiable observations at specific time steps within a socio-technological feedback-loop might even lead society to modify the desired final goal candidate(s) making it possible to ameliorate the framework as time goes by. 

		\begin{table}[h]
		\centering
		
		\begin{tabular}{ |c |c|c|c|c|c| } 
			\hline
			Ethics Framework / Focus & 	Agent  & Action  &  Outcome & Experiencer & S\&T \\ [1ex] 
			\hline
			Virtue Ethics & 	x  &  &  & & \\ [1ex] 
			\hline
			Deontological Ethics & 	 & x& & & \\ [1ex] 
			\hline
			Consequentialist Ethics (e.g.\ CU)  & 	 &  & x & & \\ [1ex] 
			\hline
			\textbf{AU} & 	\textbf{x} & \textbf{x} & \textbf{x} & \textbf{x}  & \textbf{x}\\ [1ex] 
			\hline 
			
		\end{tabular}
		\vspace{5mm}
		\caption{Decision-making focuses within different possible ethical frameworks for AGI. ``S\&T" denotes a foreseen augmentation of the ethical decision making process by science and technology including A(G)I itself. By ``experiencer", we refer to the entities in society performing the ethical evaluation via the experience of simulations (in a mental mode only or augmented).}
		\label{table:0}
	\end{table}

	\item\textit{Amalgamation of diverse perspectives}: Finally, we postulate that AU\footnote{AU is not be to confused with agent-relative consequentialism which is a normative agent-based framework, does not foresee a grounding in science and seems to assume a ``pretheoretical grasp"~\cite{schroeder2007teleology} of its ``better-than-relative-to" relation }, despite its intrinsically different motivation as a socio-technological ethical framework for AGI Safety and its non-normative nature, can be nevertheless understood as allowing a coalescence of diverse theoretical perspectives that have been historically assigned to normative ethical frameworks. To sum up and contextualize the experiencer-based AU, Table~\ref{table:0} provides an overview on the different decision-making focuses used in relevant known ethical frameworks (including CU) that might be seen as candidates for AGI Safety.

\end{itemize}

\section{Conclusion and Future Prospects}
\label{sec:conclusion}
In a nutshell, we proposed AU as a novel non-normative socio-technological ethical framework grounded in science which is conceived for the purpose of crafting ethical utility functions for AGI Safety. While CU and other classical ethical frameworks if used for AGI utility functions might engender the perverse instantiation problem, AU directly tackles this issue. AU augments CU by the following main elements: scientific grounding of utility, mental-state-dependency, debiasing of utility assignment using technology, self-reflexivity and amalgamation of diverse perspectives. Besides being able to contribute to the meaningful control of intelligent systems e.g.\ in the application domain of autonomous vehicles, AU could also be utilizable for human agents in the policy-making domain where the systems might propose ethically enhancing policy measures. Overall, we agree with Goertzel~\cite{goertzel2015superintelligence} that the perverse instantiation problem seems rather not to represent \textit{``a general point about machine intelligence or superintelligence"}. \par One of the main future challenges for the realization of AU could be the circumstance that one can only strive to approximate ethical goal functions, since a full utility elicitation on all possible future scenarios is obviously not feasible. However, already an approximation process within a socio-technological feedback-loop could lead to an ethical enhancement at a societal level. Besides that, in order to achieve safe run-time adaptive artificial intelligent systems reliably complying with ethical goal functions, a ``self-awareness" functionality might be required~\cite{aliman2018hybrid,werkhoven2018telling}. Moreover, the security of the utility function itself is essential, due to the possibility of its modification by malevolent actors during the deployment phase. In addition, it seems recommendable to require ethical goal functions to be made publicly available e.g.\ for reasons of transparency and successful AI coordination~\cite{Delphi}. Finally, proactive AGI Safety research~\cite{aliman2018hybrid}  on \textit{ethical adversarial examples} -- a conceivable type of integrity attacks on the AGI sensors having ethical consequences might be important to study in future work to complement the use of safe utility functions. 

\section*{Acknowledgements}
We would like to thank Peter Werkhoven for a helpful discussion of our approach. 

\bibliographystyle{splncs03}
\bibliography{test} 

\begin{thebibliography}{10}
\providecommand{\url}[1]{\texttt{#1}}
\providecommand{\urlprefix}{URL }

\bibitem{aliman2018hybrid}
Aliman, N.M., Kester, L.: {Hybrid Strategies Towards Safe “Self-Aware”
  Superintelligent Systems}. In: International Conference on Artificial General
  Intelligence. pp. 1--11. Springer (2018)

\bibitem{Delphi}
Aliman, N.M., Kester, L.: {Transformative AI Governance and AI-Empowered
  Ethical Enhancement Through Preemptive Simulations}. Delphi -
  Interdisciplinary review of emerging technologies  2(1),  to appear (2019)

\bibitem{arrhenius2000impossibility}
Arrhenius, G.: An impossibility theorem for welfarist axiologies. Economics \&
  Philosophy  16(2),  247--266 (2000)

\bibitem{awad2018moral}
Awad, E., Dsouza, S., Kim, R., Schulz, J., Henrich, J., Shariff, A., Bonnefon,
  J.F., Rahwan, I.: The moral machine experiment. Nature  563(7729), ~59 (2018)

\bibitem{barrett2017theory}
Barrett, L.F.: The theory of constructed emotion: an active inference account
  of interoception and categorization. Social cognitive and affective
  neuroscience  12(1),  1--23 (2017)

\bibitem{baucells2016temporal}
Baucells, M., Bellezza, S.: Temporal profiles of instant utility during
  anticipation, event, and recall. Management Science  63(3),  729--748 (2016)

\bibitem{Bentham1780-BENITT}
Bentham, J.: {An Introduction to the Principles of Morals and Legislation}.
  Dover Publications (1780)

\bibitem{berridge2014experienced}
Berridge, K.C., O'Doherty, J.P.: From experienced utility to decision utility.
  In: Neuroeconomics, pp. 335--351. Elsevier (2014)

\bibitem{bogosian2017implementation}
Bogosian, K.: Implementation of moral uncertainty in intelligent machines.
  Minds and Machines  27(4),  591--608 (2017)

\bibitem{Bostrom:2014:SPD:2678074}
Bostrom, N.: {Superintelligence: Paths, Dangers, Strategies}. Oxford University
  Press, Inc., New York, NY, USA, 1st edn. (2014)

\bibitem{busseri2011review}
Busseri, M.A., Sadava, S.W.: A review of the tripartite structure of subjective
  well-being: Implications for conceptualization, operationalization, analysis,
  and synthesis. Personality and social psychology review  15(3),  290--314
  (2011)

\bibitem{calvo2014positive}
Calvo, R.A., Peters, D.: Positive computing: technology for wellbeing and human
  potential. MIT Press (2014)

\bibitem{diener2000subjective}
Diener, E.: {Subjective well-being: The science of happiness and a proposal for
  a national index.} American psychologist  55(1), ~34 (2000)

\bibitem{diener2011happiness}
Diener, E., Biswas-Diener, R.: {Happiness: Unlocking the mysteries of
  psychological wealth}. John Wiley \& Sons (2011)

\bibitem{Eckersley2018ImpossibilityAU}
Eckersley, P.: {Impossibility and Uncertainty Theorems in AI Value Alignment
  (or why your AGI should not have a utility function)}. CoRR  abs/1901.00064
  (2018)

\bibitem{everitt2018towards}
Everitt, T.: Towards Safe Artificial General Intelligence. Ph.D. thesis, PhD
  thesis, Australian National University (2018)

\bibitem{ijcai2018-768}
Everitt, T., Lea, G., Hutter, M.: {AGI Safety Literature Review}. In:
  Proceedings of the Twenty-Seventh International Joint Conference on
  Artificial Intelligence, {IJCAI-18}. pp. 5441--5449. International Joint
  Conferences on Artificial Intelligence Organization (7 2018),
  \url{https://doi.org/10.24963/ijcai.2018/768}

\bibitem{frey2001beyond}
Frey, B.S., Stutzer, A.: {Beyond Bentham-measuring procedural utility}  (2001)

\bibitem{gilbert2007prospection}
Gilbert, D.T., Wilson, T.D.: Prospection: Experiencing the future. Science
  317(5843),  1351--1354 (2007)

\bibitem{goertzel2015superintelligence}
Goertzel, B.: Superintelligence: Fears, promises and potentials. Journal of
  Evolution and Technology  24(2),  55--87 (2015)

\bibitem{greaves2017population}
Greaves, H.: Population axiology. Philosophy Compass  12(11),  e12442 (2017)

\bibitem{johnson1994moral}
Johnson, M.: Moral imagination: Implications of cognitive science for ethics.
  University of Chicago Press (1994)

\bibitem{kahneman1999well}
Kahneman, D., Diener, E., Schwarz, N.: {Well-being: Foundations of hedonic
  psychology}. Russell Sage Foundation (1999)

\bibitem{kahneman1997back}
Kahneman, D., Wakker, P.P., Sarin, R.: Back to bentham? explorations of
  experienced utility. The quarterly journal of economics  112(2),  375--406
  (1997)

\bibitem{kaminitz2019contemporary}
Kaminitz, S.C.: {Contemporary Procedural Utility and Hume's Early Idea of
  Utility}. Journal of Happiness Studies pp. 1--14 (2019)

\bibitem{kaufman2018self}
Kaufman, S.B.: {Self-Actualizing People in the 21st Century: Integration With
  Contemporary Theory and Research on Personality and Well-Being}. Journal of
  Humanistic Psychology p. 0022167818809187 (2018)

\bibitem{koltko2006rediscovering}
Koltko-Rivera, M.E.: {Rediscovering the later version of Maslow's hierarchy of
  needs: Self-transcendence and opportunities for theory, research, and
  unification}. Review of general psychology  10(4),  302--317 (2006)

\bibitem{van2018virtual}
van Loon, A., Bailenson, J., Zaki, J., Bostick, J., Willer, R.: Virtual reality
  perspective-taking increases cognitive empathy for specific others. PloS one
  13(8),  e0202442 (2018)

\bibitem{lopez2018positive}
Lopez, S.J., Pedrotti, J.T., Snyder, C.R.: {Positive psychology: The scientific
  and practical explorations of human strengths}. Sage Publications (2018)

\bibitem{lyubomirsky2001some}
Lyubomirsky, S.: {Why are some people happier than others? The role of
  cognitive and motivational processes in well-being.} American psychologist
  56(3),  239 (2001)

\bibitem{maslow1971farther}
Maslow, A.H.: The farther reaches of human nature. Viking (1971)

\bibitem{meuhlhauser2012intelligence}
Meuhlhauser, L., Helm, L.: {Intelligence Explosion and Machine Ethics}.
  Singularity Hypotheses: A Scientific and Philosophical Assessment pp.
  101--126 (2012)

\bibitem{mossbridge2018emotionally}
Mossbridge, J., Goertzel, B., Mayet, R., Monroe, E., Nehat, G., Hanson, D., Yu,
  G.: {Emotionally-sensitive AI-driven android interactions improve social
  welfare through helping people access self-transcendent states. vol}. In: AI
  for Social Good Workshop at Neural Information Processing Systems 2018
  Conference (2018)

\bibitem{oosterwijk2012states}
Oosterwijk, S., Lindquist, K.A., Anderson, E., Dautoff, R., Moriguchi, Y.,
  Barrett, L.F.: States of mind: Emotions, body feelings, and thoughts share
  distributed neural networks. NeuroImage  62(3),  2110--2128 (2012)

\bibitem{parfit1984reasons}
Parfit, D.: Reasons and persons. OUP Oxford (1984)

\bibitem{peterson2006primer}
Peterson, C.: A primer in positive psychology. Oxford University Press (2006)

\bibitem{peterson2010royal}
Peterson, M.: A royal road to consequentialism? Ethical theory and moral
  practice  13(2),  153--169 (2010)

\bibitem{schroeder2007teleology}
Schroeder, M.: Teleology, agent-relative value, and ‘good’. Ethics  117(2),
   265--295 (2007)

\bibitem{seligman2012flourish}
Seligman, M.E.: Flourish: A visionary new understanding of happiness and
  well-being. Simon and Schuster (2012)

\bibitem{seligman2014positive}
Seligman, M.E., Csikszentmihalyi, M.: {Positive psychology: An introduction}.
  In: Flow and the foundations of positive psychology, pp. 279--298. Springer
  (2014)

\bibitem{werkhoven2018telling}
Werkhoven, P., Kester, L., Neerincx, M.: Telling autonomous systems what to do.
  In: Proceedings of the 36th European Conference on Cognitive Ergonomics.
  p.~2. ACM (2018)

\bibitem{y}
Yampolskiy, R.V.: Utility function security in artificially intelligent agents.
  Journal of Experimental \& Theoretical Artificial Intelligence  26(3),
  373--389 (2014)

\bibitem{yampolskiy2015artificial}
Yampolskiy, R.V.: Artificial Superintelligence: A Futuristic Approach. Chapman
  and Hall/CRC (2015)

\bibitem{yampolskiy2019personal}
Yampolskiy, R.V.: {Personal Universes: A Solution to the Multi-Agent Value
  Alignment Problem}. arXiv preprint arXiv:1901.01851  (2019)

\bibitem{zie}
Ziesche, S.: {Potential Synergies Between The United Nations Sustainable
  Development Goals And The Value Loading Problem In Artificial Intelligence}.
  Maldives National Journal of Research  6, ~47 (06 2018)

\end{thebibliography}

\end{document}